\documentclass[letterpaper]{article}
\usepackage[preprint]{nips_2018}
\usepackage{graphicx}
\usepackage{subfigure}

\usepackage[utf8]{inputenc} 
\usepackage[T1]{fontenc}    
\usepackage{hyperref}       
\usepackage{url}            
\usepackage{booktabs}       
\usepackage{amsfonts}       
\usepackage{nicefrac}       
\usepackage{microtype}      
\usepackage{amsmath,amscd,amssymb}
\usepackage{wrapfig}
\usepackage{listings}

\usepackage{times}

\title{Probabilistic Meta-Representations Of Neural Networks}

\author{
  \textbf{Theofanis Karaletsos}$^1$,\ \textbf{Peter Dayan}$^{1,2}$, \\
  \textbf{Zoubin Ghahramani}$^{1,3}$\\
  $^1$ Uber AI Labs, San Francisco, CA, USA\\
  $^2$ University College London, United Kingdom \\
  $^3$ Department of Engineering, University of Cambridge, United Kingdom \\
  \texttt{\href{mailto:theofanis@uber.com}{theofanis@uber.com}}, \texttt{\href{dayan@gatsby.ucl.ac.uk}{dayan@gatsby.ucl.ac.uk}}, \\
  \texttt{\href{mailto:zoubin@eng.cam.ac.uk}{zoubin@eng.cam.ac.uk}}
}

\begin{document}
\maketitle

\begin{abstract}
Existing Bayesian treatments of neural networks are typically characterized by weak prior and approximate
posterior distributions according to which all the weights are drawn
independently. Here, we consider a richer prior distribution in which units in the
network are represented by latent variables, and the weights between
units are drawn conditionally on the values of the collection of those variables. This allows rich correlations between related weights, and can be seen as realizing a function prior with a Bayesian complexity
regularizer ensuring simple solutions. We illustrate the resulting
meta-representations and representations, elucidating the power of this
prior.
\end{abstract}

\section{Introduction}
\vspace{-0.1in}

Neural networks are ubiquitous model classes in machine
learning. Their appealing computational properties, intuitive
compositional structure and universal function
approximation capacity make them a straightforward choice for tasks requiring complex function mappings. 

In such tasks, there is an inevitable battle between flexibility and generalization. Two very different sorts of weapon are popular: one is to incorporate neural networks into the canon of Bayesian inference, using prior distributions over weights to regularize the mapping. However, historically, very simple prior distributions over the weights have been used which are based more on computational convenience than appropriate model specification. Concomitantly, simple approximations to posterior distrbutions are employed, for instance failing to capture the correlations between weights. These frequently fail to to cope with the richness of the underlying inference problem. The second weapon is to reduce the number of effective parameters, typically by sharing weights (as in convolutional neural nets; CNNs). In CNNs, units are endowed with (topographic) locations in a geometric space of inputs, and the weights between units are made to depend systematically (though heterogeneously) on these locations. However, this solution is minutely specific to particular sorts of mapping and task.

Here, we combine these two arms by proposing a higher-level abstraction in which the units of the network themselves are probabilistically embedded into a shared, structured, meta-representational space (generalizing topographic location), with weights and biases being derived conditional on these embeddings. Rich structural patterns can thus be induced into the weight distributions. Our model captures uncertainty on three levels:
meta-representational uncertainty, function uncertainty given the embeddings, and observation
(or irreducible output) uncertainty.  This hierarchical decomposition is flexible,
and is broadly applicable to modeling task-appropriate weight priors,
weight-correlations, and weight uncertainty. It can also be beneficially
used in the context of various modern applications, where the ability to perform structured weight manipulations online is beneficial.

We first describe probabilistic and Bayesian neural networks in general in Sec.~\ref{sec:probnn}. We then describe our meta-representation in Sec.~\ref{sec:meta-intro} and propose its use as meta-prior in a generative model of weights in Sec.~\ref{sec:meta-model}. We show a set of indicative experiments in Sec.~\ref{sec:exp}, and discuss related work in Sec.~\ref{sec:related_work}.

\section{Probabilistic Neural Networks}
\label{sec:probnn}

Let $\mathcal{D}$ be a dataset of $n$ tuples $\{ (x_1, y_1), ..., (x_n,
y_n ) \}$ where $x$ are inputs and $y$ are targets for supervised
learning. Take a neural network (NN) with $L$ layers, 
$V_{l}$ units in layer $l$ (we drop $l$ where this is clear),  an overall collection of weights and biases
$\mathcal{W}=\{{\bf W}_l\}_{1:L}$, and fixed nonlinear activation functions. In Bayesian terms, the NN realizes
the likelihood $p(y|x,\mathcal{W})$; together with a
prior $P(\mathcal{W})$ over $\mathcal{W}$, this generates the
conditional distribution (also known as the marginal likelihood) 
$P({\bf y}|{\bf x}) = \int P({\bf y}|{\bf x}, \mathcal{W}) P(\mathcal{ W}) d\mathcal{W}$,
where $\bf{x}$ denotes $(x_1, \ldots, x_n)$ and $\bf{y}$ denotes $(y_1, \ldots, y_n)$.
One common assumption for the prior is that the weights are drawn iid
from a zero-mean common-variance normal distribution, leading to a prior which factorizes across layers and units:
\begin{equation*}
\begin{split}
P(\mathcal{W}) &= \prod_{l=1}^{L}\prod_{i=1}^{V_l}\prod_{j=1}^{V_{l-1}} P(w_{l,i,j}) \\&= \prod_{l=1}^{L}\prod_{i=1}^{V_l}\prod_{j=1}^{V_{l-1}} \mathcal{N}(w_{l,i,j}|0,\lambda),
\end{split}
\end{equation*}
where $i$ and $j$ index units in adjacent layers of the network, and $\lambda$ is the prior weight variance.  Maximum-a-posteriori inference with this prior famously results in an objective identical to  $L_2$ weight regularization with regularization constant $1/\lambda$.  Other suggestions for priors
have been made, such as \citep{neal1996priors}'s proposal that the
precision of the prior should be scaled according to the number of
hidden units in a layer, yielding a contribution to the prior of
$\mathcal{N}(0,\frac{1}{V_l})$.

Bayesian learning requires us to infer the posterior distribution of
the weights given the data $\mathcal{D}$:
$P(\mathcal{W} | \mathcal{D}) = P(\mathcal{W}) \prod_{i=1}^{n} P(y_i|x_i, \mathcal{W})/P({\bf y}|{\bf x})$. 
Unfortunately, the marginal likelihood and posterior are intractable as
they involve integrating over a high-dimensional space defined by the
weight priors.
A common step is therefore to perform approximate inference, for
instance by varying the parameters $\Phi$ of an approximating distribution $Q( \mathcal{W} ; \Phi)$ to make it close to the true posterior. For instance, in Mean Field
Variational Inference ({\bf MF-VI}), we consider the factorized posterior:
\begin{equation*}
Q( \mathcal{W} ; \Phi) = \prod_{l=1}^{L}\prod_{i=1}^{V_l}\prod_{j=1}^{V_{l-1}} Q(w_{l,i,j} ; \pmb{\phi}_{l,i,j}).
\end{equation*}
Commonly, $Q$ is Gaussian
for each weight, $Q(w_{l,i,j} ; \pmb{\phi}_{l,i,j})= \mathcal{N}({\bf w}_{l,i,j} |
\mu_{l,i,j}, \sigma^{2}_{l,i,j})$, with variational parameters $\pmb{\phi}_{l,i,j} = \{ \mu_{l,i,j},
\sigma^{2}_{l,i,j} \}$. The parameters $\Phi$ are adjusted to maximize a lower bound $\mathcal{L}(\Phi)$ to the marginal likelihood given by the Evidence Lower Bound ({\bf ELBO}) (a relative of the free energy):
\begin{equation}
\label{eq:elbo}
\mathcal{L}(\Phi) = \mathbb{E}_{Q(\mathcal{W})} \big [ \text{log}P({\bf y}|{\bf x}, \mathcal{W}) + \text{log} P(\mathcal{W})\\ - \text{log}Q(\mathcal{W}) \big ].
\end{equation}
The mean field factorization assumption renders maximizing the {\bf ELBO} tractable for many models. The predictive distribution of a Bayesian Neural Network can be approximated utilizing a mixture distribution $P(y| x, \mathcal{D}) \approx \frac{1}{S} \sum\nolimits_{s=1}^{S}  P(y| x, \mathcal{W}^{s})$ over $S$ sampled instances of weights $\mathcal{W}^{s} \sim Q(\mathcal{W})$.

\section{Meta-Representations of Units}
\label{sec:meta-intro}
We suggest abandoning \emph{direct} characterizations of weights or distributions over
weights, in which weights are individually independently tunable. Instead, we couple weights using \emph{meta-representations} (so called, since they determine the parameters of the underlying NN that
themselves govern the input-output function represented by the NN). These treat the units as the
primary objects of interest and embed them into a shared space, deriving weights as secondary structures.

 Consider a code ${\bf z}_{l,u} \in \Re^D$ that uniquely describes each unit $u$ 
(visible or hidden) in layer $l$ in the network.  Such codes could for example be one-hot codes or Euclidean embeddings of units in a real space $\Re^K$. A generalization is to use an inferred \emph{latent representation} which embeds units $l,u$ in a $D$-dimensional vector space. 
Note that this code encodes the unit itself, not its activation.

Weighs $w_{l,i,j}$ linking two units can then be recast in terms of those units'
codes ${\bf z}_{l,i}$ and ${\bf z}_{l-1,j}$ for instance by concatenation ${\bf z}_{w}(l,i,j) =
\big[{\bf z}_{l,i}, {\bf z}_{l-1,j}\big]$. We call the collection of all
such weight codes  $\mathcal{Z}_w$ (which can be deterministically derived from the collection of unit codes ${\bf Z}$). Biases ${\bf b}$ can be constructed similarly, for instance using $0$'s as the second code; thus we do not distinguish them below. Weight codes then form a conditional prior distribution 
$P(w_{l,i,j}| {\bf z}_{w}(l,i,j))$, parameterized by a function $g({\bf z}_{w}(l,i,j), \xi)$ shared across the entire network. Function $g$, which may itself have parameters $\xi$, acts as a conditional hyperprior that gives rise to a prior over the weights of the
original network:
\begin{equation*}
P(\mathcal{W} | \mathcal{Z}_w; \xi) \!=\!
\prod_{l=1}^{L} \prod_{i=1}^{V_l} \prod_{j=1}^{V_{l-1}}\! P(w_{l,i,j}|{\bf z}_{w}(l,i,j) ;\xi).
\end{equation*}

There remain various choices: we commonly use either Gaussian or implicit observation models for the weight prior and neural networks as hyperpriors (though clustering and Gaussian Processes merit exploration). Further, the  weight code can be augmented with a global state variable ${\bf z_s}$ (making
${\bf z}_{w}(l,i,j)=\big[{\bf z}_{l,i},{\bf z}_{l-1,j},{\bf z_s}\big]$) 
which can  coordinate all the weights or add conditioning knowledge. 

Examples for observation models for weights is a Gaussian model:
\begin{equation*}
\big[ \mu_{w(l,i,j)}, \Sigma_{w(l,i,j)}  \big] = g({\bf z}_{w}(l,i,j), \xi)
\end{equation*}
and similarly an implicit model:
\begin{equation*}
\big[ \mu_{w(l,i,j)}  \big] = g({\bf z}_{w}(l,i,j), \epsilon, \xi), 
\end{equation*}
with$ \epsilon \sim \mathcal{N}(0,1)$, which produces arbitrary output distributions.

\section{MetaPrior: A Generative Meta-Model of Neural Network Weights}
\label{sec:meta-model}
\vspace{-0.1in}
\begin{wrapfigure}{R}{0.28\textwidth}
    \includegraphics[width=0.28\textwidth]{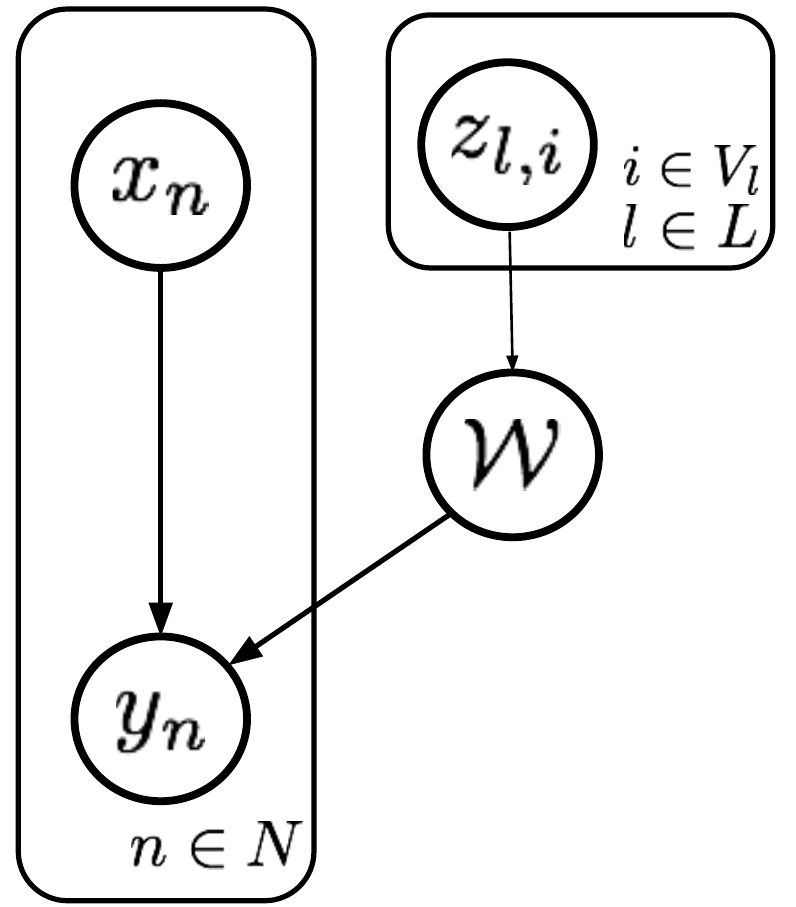}
   \caption{A graphical model showing the MetaPrior. Note the additional plate over the meta-variables {\bf Z} indicating the distribution over meta-representations. }
\label{fig:graphical_model}
\end{wrapfigure}

The meta-representation can be used as a prior for Bayesian NNs. Consider a case with latent codes for units being sampled from $P({\bf Z}) = \prod_{l,i} P({\bf z}_{l,i})$, and  the weights of the underlying NN being sampled according to the conditional distribution for the weights $P(\mathcal{W}|\mathcal{Z}_w) $. Put together this yields the model:
\begin{equation*}
\begin{split}
&P({\bf y}|{\bf x}) = \int \nolimits_{{\bf Z}} P({\bf Z}) \int \nolimits_{\mathcal{W}} P({\bf y}|{\bf x}, \mathcal{W}) P(\mathcal{W}|{\bf Z}) d\mathcal{W}d{\bf Z} \label{eq:metap} 
\\&P({\bf Z})= \prod\nolimits_{l,i}P({\bf z}_{l,i}) = \prod_{l,i} \mathcal{N}({\bf 0},{\bf 1}).
\end{split}
\end{equation*}
Crucially, the conditional distribution over weights depends on more than one unit representation. This can be seen as a structural form of weight-sharing or a function prior and is made more explicit using the plate notation in Fig~\ref{fig:graphical_model}.
Conditioned on a set of sampled variables {\bf Z}, our model defines a particular space of functions  $f_{{\bf Z}}: {\bf X} \rightarrow {\bf Y}$. 
We can recast the predictive distribution given training data $\mathcal{D^{*}}$ as:
\begin{equation*}
P({\bf y}|{\bf x},\mathcal{D^{*}}) = \int \limits_{{\bf Z}} Q({\bf Z}|\mathcal{D^{*}}) \int \limits_{f_{{\bf Z}}} P({\bf y}|{\bf x}, f_{{\bf Z}}) P(f_{{\bf Z}}|{\bf Z}) df_{{\bf Z}} d{\bf Z}.
\end{equation*}
Uncertainty about the unit embeddings affords the model the flexibility to represent diverse functions by coupling weight distributions with meta-variables.

The learning task for training MetaPriors for a particular dataset $\mathcal{D}$ consists of inferring the posterior distribution of the latent variables $P({\bf Z}|\mathcal{D})$.
Compared with the typical training loop in Bayesian Neural Networks, which involves learning a posterior distribution over weights, the posterior distribution that needs to be inferred for MetaPriors is the approximate distribution $Q({\bf Z} ;\Phi)$ over the collection of unit variables as the model builds meta-models of neural networks.
We train by maximizing the evidence lower bound ({\bf ELBO}):
\begin{equation}
\text{log}P(\mathcal{D};\xi) \geq \mathbb{E}_{Q(\bf Z)}\Big[\text{log}\frac{ P(\mathcal{D}|{\bf Z} ;\xi) P({\bf Z}) }{Q({\bf Z};\Phi)}\Big]
\end{equation}
with $\text{log}P(\mathcal{D}|{\bf Z}) \approx \frac{1}{S}\sum \limits_{s=1}^{S} P({\bf y}|{\bf x},\mathcal{W}^{s,m})$ and $\mathcal{W}^{s,m} \sim P(\mathcal{W}|{\bf Z}^{m})$.

In practice, we apply the reparametrization trick~\citep{kingma2014stochastic, rezende2014stochastic, titsias2015local} and its variants~\citep{kingma2015variational} and subsample ${\bf Z}^{m} \sim Q({\bf Z};\Phi)$ to maximize the objective:
\begin{equation}
\label{eq:svi_meta}
\begin{split}
\text{{\bf ELBO}}(\Phi, \xi) = &\frac{1}{M}\sum \limits_{m=1}^{M} \Big[ \frac{1}{S}\sum \limits_{s=1}^{S} \text{log}P({\bf y}|{\bf x}, \mathcal{W}^{s,m}) \Big] \\&- \text{KL}\big(Q({\bf Z};\Phi)||P({\bf Z})\big).
\end{split}
\end{equation}

However, pure posterior inference over $\Phi$ without gradient steps to update $\xi$ results in learning meta-representations which best explain the data given the current hypernetwork with parameters $\xi$. We call the process of inferring representations $P({\bf Z}|\mathcal{D})$ {\it illation}. Intuitively, inferring the meta-representations for a dataset induces a functional alignment to its input-output pairs and thus reduces variance in the marginal representations for a particular collection of data points.

We can also recast this as a two-stage learning procedure, similar to Expectation Maximization (EM), if we want to update the function parameters $\xi$ and the meta-variables independently.
First, we approximate $P(\bf Z|\mathcal{D};\xi)$ by $Q({\bf Z}; \phi)$ by maximizing
$\text{{\bf ELBO}}(\phi ; \xi)$. Then, we can maximize of $\text{{\bf ELBO}}(\xi;\phi)$ to update the hyperprior function. In practice, we find that Eq.~\ref{eq:svi_meta} performs well with a small amount of samples for learning, but using EM can help reduce gradient variance in the small-data setting. 

We highlight that a particularly appealing property of this {\bf ELBO} is that the weight observation term $P(\mathcal{W}|{\bf Z})$ appears in both the model and the variational approximation and as such analytically cancels out. We exploit this property in that we can use implicit weight models without having to resort to using a discriminator as is typically the case when using GAN-style inference.

\section{Experiments}
\label{sec:exp}
\vspace{-0.1in}
We illustrate the properties of MetaPriors with a series of tasks of increasing complexity, starting with simple regression and classification, and then graduating to few-shot learning. 
\vspace{-0.1in}
\subsection{Toy Example: Regression}
\label{sec:regression}
\vspace{-0.1in}
Here, we consider the toy regression task popularized in~\citep{hernandez2015probabilistic} ( $y = x^{3} + \epsilon$ with $\epsilon \sim \mathcal{N}(0,3)$). We use neural networks with a fixed observation noise given by the model, and seek to learn suitably uncertain functions.
For all networks, we use 100 hidden units, and 2-dimensional latent codes and 32 hidden units in the hyper-prior network $f$.

\begin{figure}[h]
\begin{center}
\includegraphics[width=0.9\linewidth]{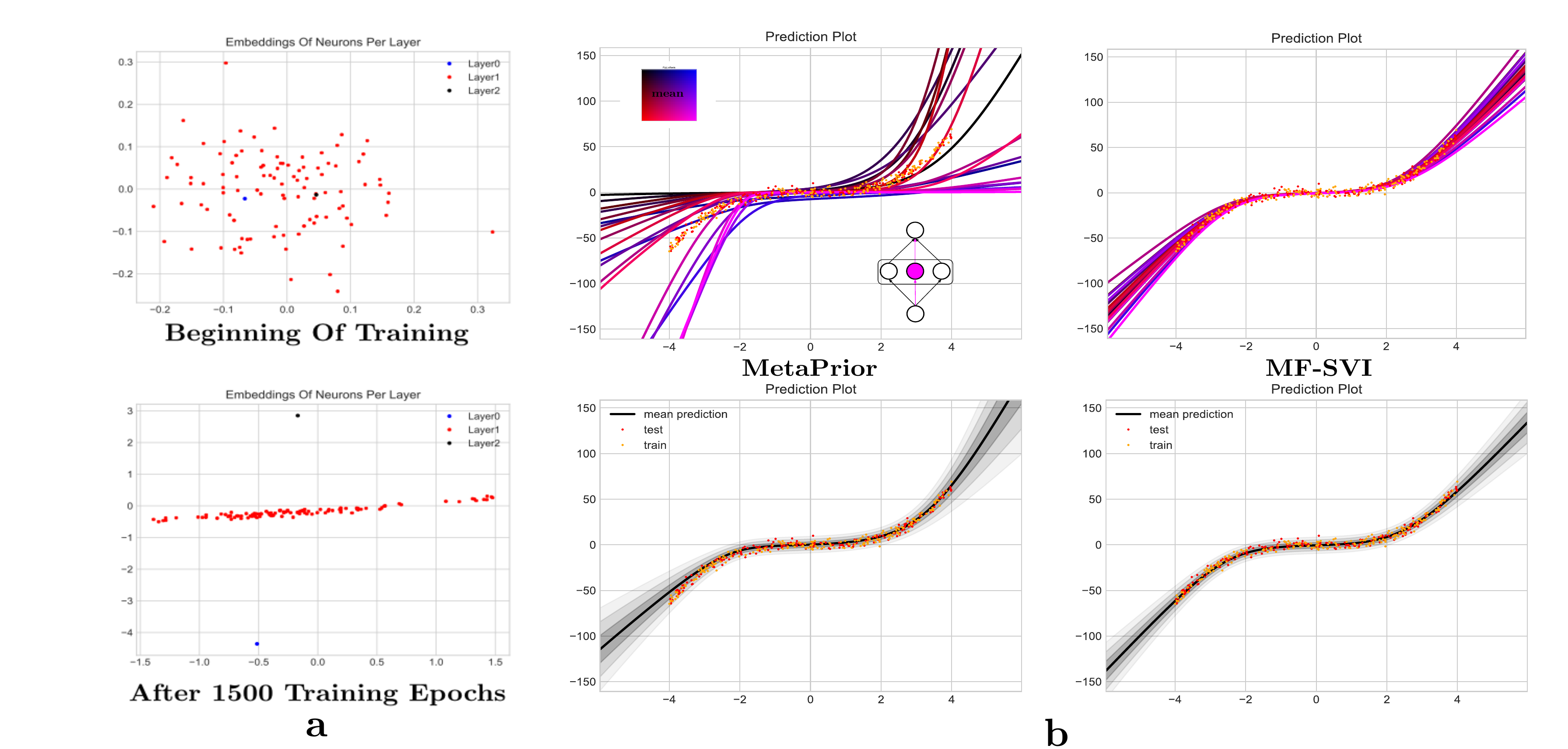}
\end{center}
\caption{{\bf MetaPrior as a Function Prior}: {\bf (a)} Means of the meta-variables ${\bf Z}$ of MetaPrior model embedded in shared 2d space at the onset of training and at the end show how the variables perform a structured topographic mapping over the embedding space. {\bf (b)} ~{\bf Left}: MetaPrior fit and function draws $f_{\bf Z}$ are visualized. Functions are drawn by keeping all meta-variables clamped to the mean except a random one among the hidden layer units which is then perturbed according to the offset indicated by the color-legend. Changes in a single meta-variable induce global changes across the entire function. The function space itself is interesting as the model appears to have generalized to a variety of smoothly varying functions as we change the sampling offset, the mean of which is the cubic function. This is a natural task for our model, as all units of the hidden layer are embedded in the same space and sampling means exploring that space and all the functions this induces.
 {\bf Right}: A {\bf MF-VI} BNN with fixed observation noise function fit to the toy example and function draws are shown. The function draws are performed by picking 40 weights at random and sampling from their priors, while keeping the rest clamped to their posterior means. Single weight changes induce barely perceptible function fluctuations and in order to explore the representational ability of the model we sample multiple weights at once.}
\label{fig:toy_regression}
\end{figure}

In Fig.~\ref{fig:toy_regression} we show two function fits to this example: our model and a mean field network.
We observe that both models increase uncertainty away from the data, as would be expected. 
We also illustrate function draws which show how each model uses its respective weight representation differently to encode functions and uncertainty.
We sample functions in both cases by clamping latent variables to their posterior mean and allowing a single latent variable to be sampled from its prior.
Our model, the MetaPrior-based Bayesian NN, learns a global form of function uncertainty and has global diversity over sampled functions for sampling just a single latent variable.
It uses the composition of these functions through the meta-representational uncertainty to model the function space.
This can be attributed to the strong complexity control enacted by the pull of the posterior fitting mechanism on the meta-variables. Maximum a posteriori fits of this model yielded just the mean line directly fitting the data.
The mean field example shows dramatically less diversity in function samples, we were forced to sample a large amount of weights to elicit the diversity we got as single weight samples only induced small local changes.
This suggests that the MetaPrior may be capturing interesting properties of the weights beyond what the mean field approximation does, as single variable perturbations have much more impact on the function space.

\begin{figure}[h!]
\begin{center}
\includegraphics[width=1.0\linewidth]{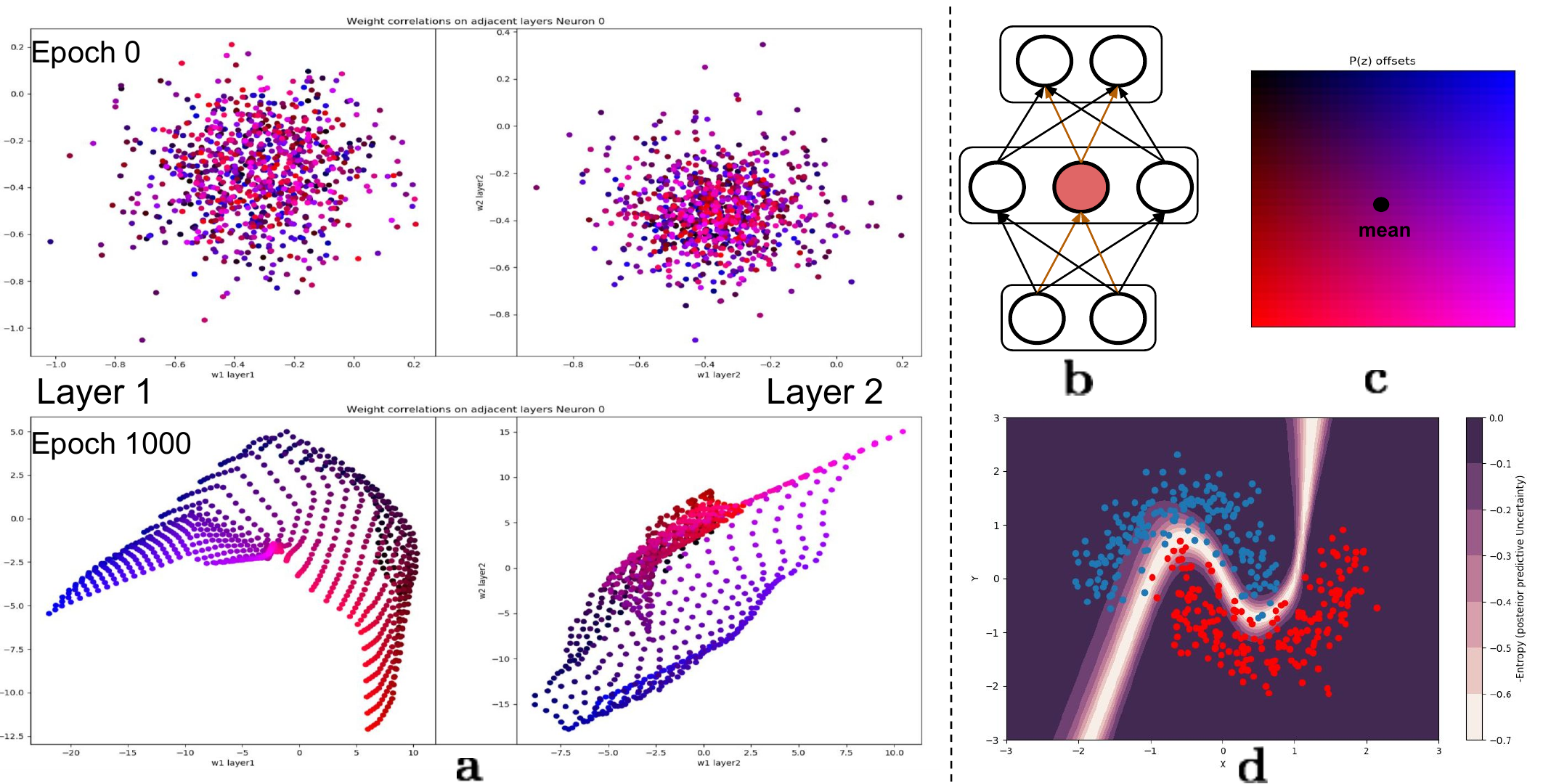}
\end{center}
\caption{ {\bf Classification and Weight Correlation Example}:~{\bf (a)} We illustrate correlation structures in the marginal weight distributions for the Half Moon classification example. All units ${\bf Z}$ are clamped to their mean. We sample a single unit ${\bf z}_0$ repeatedly from its 2-dimensional distribution and sample $P(\mathcal{W}|{\bf Z})$. We track how particular ancestral samples generate weights by color-coding the resulting sampled weights by the offset of the ancestral sample from its mean. Similar colors indicate similar samples. Horizontal axes denote value for weight 1, vertical weight for 2, so that each point describes a state of a weight column in a layer. {\bf Left}: We show the sampled weights from layer 1 of the neural network connected to unit 0. {\bf Right}: The sampled weights from the output layer 2 connected to unit 0. {\bf Top}: At the onset of training the effect of varying the unit meta-variable ${\bf z}_u$ is not observable. {\bf Bottom}: After learning the function shown in SubFig.~{\bf d} we can see that the effect of varying the unit meta-variable induces correlations within each layer and also across both layers.
{\bf (b)} Sketch of the NN  used with highlighted unit whose meta-variable we perturb and the affected weights.
{\bf (c)} Legend encoding shifts from the mean in the meta-variable to colors.
{\bf (d)} Predictive uncertainty of the classifier and the predicted labels of the datapoints, demonstrating that the model learns more than a point estimate for the class boundary.}
\label{fig:weight_corr}
\end{figure}

\vspace{-0.1in}
\subsection{Toy Example: Classification}
\vspace{-0.1in}
We illustrate the model's function fit to the half-moon two class classification task in Fig.~\ref{fig:weight_corr}, also visualizing the learned weight correlations by sampling from the representation.
The model reaches 95.5\% accuracy, on par with a mean field BNN and an MLP.
Interestingly, meta-representations  induce intra- and inter-layer correlations of weights, amounting to a form of soft weight-sharing with long-range correlations. This visualizes the mechanisms by which complex function draws as observed in Sec.~\ref{sec:regression} are feasible with only a single variable changing. 
The model captures structured weight correlations which enable global weight changes subject to a low-dimensional parametrization. This is a conceptual difference to networks with individually tunable weights.

\vspace{-0.1in}
\subsection{MNIST50k-Classification}
\label{sec:mnist_class}
\begin{figure}[h!]
\begin{center}
\includegraphics[width=1.0\linewidth]{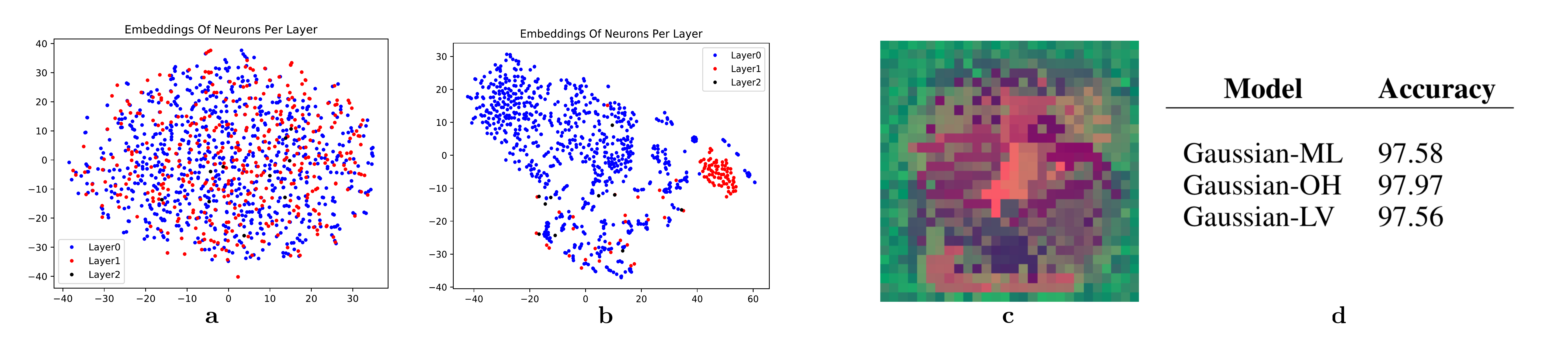}
\end{center}
\caption{ {\bf MNIST Study}: {\bf (a)} T-SNE visualization of the units in an MNIST network before training reveals random structure of the units of various layers.  {\bf (b)} Training induces structured embeddings of input, hidden and class units. 
{\bf (c)} Input units color coded by coordinates of corresponding unit embeddings are visualized. 
Interestingly, many units in the input layer appear to cluster according to their marginal occupancy of digits. This is an expected property for the boundary pixels in MNIST which are always empty.
The model represents those input pixels with similar latent units $Z$ and effectively compresses the weight space.
{\bf (d)} Performance table of MNIST models.}
\label{fig:mnist_embedding}
\end{figure}

\vspace{-0.1in}
We use NNs with one hidden layer and 100 hidden units to test the simplest model possible for MNIST-classification. We train and compare the deterministic one-hot embeddings (Gaussian-OH), with the latent variable embeddings  (Gaussian-LV) used elsewhere (with 10 latent dimensions); along with mean field NNs with a unit Gaussian prior (Gaussian-ML). 
We  visualize the learned ${\bf Z}$s (Fig.~\ref{fig:mnist_embedding}) by producing a T-SNE embedding of their means which reveal relationships among the units in the network. The figure shows that the model infers semantic structure in the input units, as it compresses boundary units to a similar value. This is representationally efficient as no capacity is wasted on modeling empty units repeatedly.

\begin{figure}[h!]
\begin{center}
\includegraphics[width=1.0\linewidth]{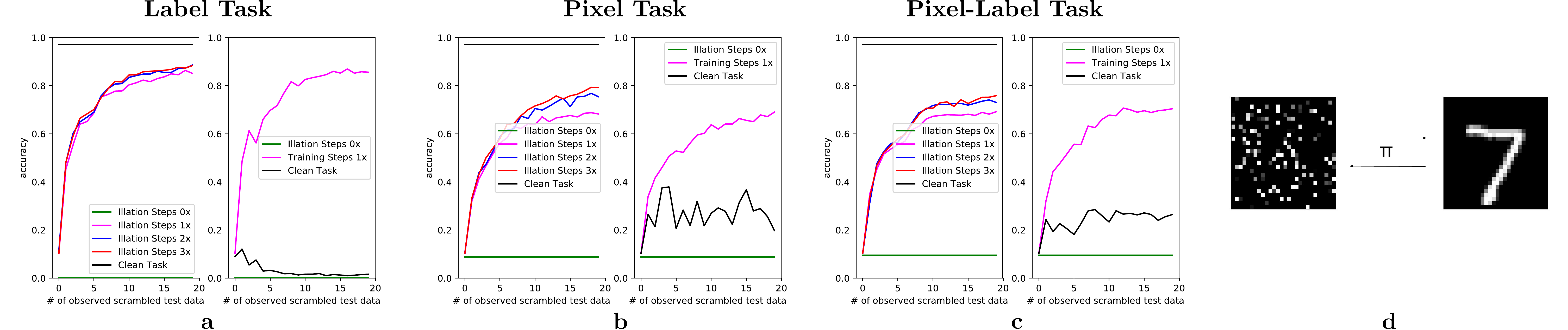}
\end{center}
\caption{{\bf Few-shot Disentangling Task} {\bf (a)-(c)}~{\bf Left}: Performance of MetaPrior when inferring $Q({\bf Z}|\mathcal{D})$ as a function of inference work (color coding) and revealed test data (shots divided by 20) for adaptation to a new task. {\bf Right}: Performance of the baseline MF-BNN model decays on the clean task. {\bf (d)} Illustration of the pixel permutation we stress the model with.}
\label{fig:few_shot}
\end{figure}
\vspace{-0.1in}
\subsection{Few-Shot and Multi-Task Learning}
\label{sec:multi-task}
\vspace{-0.1in}
The model allows task-specific latent variables ${\bf Z}_t$ to be used to learn representations for separate, but related, tasks, effectively ${\it aligning}$ the model to the current input data.
In particular, the hierarchical structure of the model facilitates few-shot learning on a new task $t$ by inferring $P({\bf Z}_t|\mathcal{D}_t)$, starting from the ${\bf Z}$ produced by the original training, and keeping fixed the mapping $P(\mathcal{W}|{\bf Z}_t ;\xi)$.
We tested this ability using the MNIST network. After learning the previous model for clean MNIST, we evaluate the model's performance on the MNIST test data in both a clean and a permuted version. 
We sample random permutations ${\bf \pi}$ for input pixels and classes and apply them to the test data, visualized in Fig.~\ref{fig:few_shot}. 
This yields three datasets, with permutations of input, output or both.
We then proceed to illate $P({\bf Z}_t|\mathcal{D}_t)$ on progressively growing numbers of scrambled observations (shots) $\mathcal{D}_t$. For each such attempt, we reset the model's representations to the original ones from clean training. As a baseline, we train a mean field neural network. We also keep track of performance on the clean dataset as reported in Fig.~\ref{fig:few_shot}.
We examine a related setting of generalization in the Appendix Sec.~\ref{sec:surgery}.

\begin{figure}[h!]
\begin{center}
\includegraphics[width=1.0\linewidth]{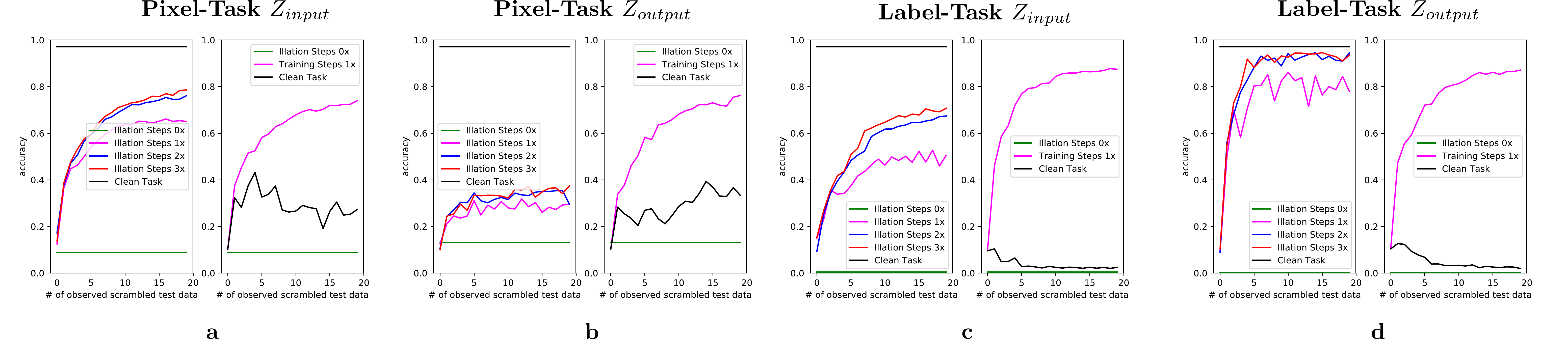}
\end{center}
\caption{{\bf Structured Surgery Task} {\bf (a)-(d)}~{\bf Left}: Performance of MetaPrior when inferring $Q({\bf Z}_{\cdot}|\mathcal{D})$ as a function of inference work (color coding) and revealed test data (shots divided by 20) for adaptation to a new task. {\bf Right}: Performance of the baseline MF-BNN model decays on the clean task.}
\label{fig:surgery}
\end{figure}

\section{Structured Surgery}
\label{sec:surgery}
In the multi-task experiments in Sec.~\ref{sec:multi-task}, we studied the model's ability to update representations holistically and generalize to unseen variations in the training data.
Here, we manipulate the meta-variables in a structured and targeted way (see Fig.~\ref{fig:surgery}).
Since ${\bf Z}= \{{\bf Z}_{input},{\bf Z}_{hidden},{\bf Z}_{output} \}$, we can elect to perform illation only on a subset of variables.
Instead of learning a new set of task-dependent ${\bf Z}_{t}$ variables, we only update input or output variables per task to demonstrate how the model can disentangle the representations it models and can generalize in highly structured fashion.
When updating only the input variables, the model reasons about pixel transformations, as it can move the representation of each input pixel around its latent feature space.
The model appears to solve for an input permutation by searching in representation space for a program approximating ${\bf \tilde{Z}}_{input} = {\bf \pi}({\bf Z}_{input})$. This search is ambiguous, given little data and the sparsity underlying MNIST. This process demonstrates the alignment the model automatically performs only of its inputs in order to change the weight distribution it applies to datasets, while keeping the rest of the features intact.
Similarly, we observe that updating only the class-label units leads to rapid and effective generalization for class shifts or, in this case, class permutation, since only 10 variables need to be updated. The model could also easily generalize to new subclasses smoothly existing between current classes.
These demonstrate the ability of the model to react in differentiated ways to shifts, either by adapting to changes in input-geometry, target semantics or actual features in a targeted way while keeping the rest of the representations constant.

\section{Related Work}
\label{sec:related_work}
Our work brings together two themes in the literature. One is the probabilistic interpretation of weights and activations of neural networks, which has been a common approach to regularization and complexity control~\citep{ mackay1992bayesian, mackay1992practical, mackay1995bayesian, hinton1993keeping, dayan1995helmholtz, hinton1995wake, neal2012bayesian,blundell2015weight, hernandez2015probabilistic}. The second theme is to consider the structure and weights of neural networks as arising from embeddings in other spaces. This idea has been explored in evolutionary computation~\citep{stanley2007compositional, gauci2010autonomous, risi2012enhanced, clune2011performance} and beyond, and applied to recurrent and convolutional NNs and more. Our learned hierarchical probabilistic representation of units, which we call a meta-representation because of the way it generates the weights, is inspired by this work. It can thus also be considered as a richly structured hierarchical Bayesian Neural network~\citep{finkel2009hierarchical,joshi2016hierarchical}.
In important recent work training ensembles of neural
networks~\citep{lakshminarayanan2017simple} was proposed. This captures uncertainty
well; but ensembles are a departure from a single, self-contained model.

Our work is most closely related to two sets of recent studies. One considers reweighting activation patterns to improve posterior inference~\citep{krueger2017bayesian, pawlowski2017implicit, louizos2017multiplicative}. The use of parametric weights and normalizing flows~\citep{rezende2015variational, papamakarios2017masked, dinh2016density} to model scalar changes to those weights offers a probabilistic patina around forms of batch normalization. 
However, our work is not aimed at capturing posterior uncertainty for given weight priors, but rather as a novel weight prior in its own right.
Our method is also representationally more flexible, as it provides embeddings for the weights as a whole. 

Equally, our meta-representations of units is reminiscent of the inducing points that are used to simplify Gaussian Process (GP) inference \citep{quinonero2005unifying, snelson2006sparse, titsias2009variational}, and that are key components in GP latent variable models~\citep{lawrence2004gaussian, titsias2010bayesian}. Rather like inducing points, our units control the modeled function and regularize its complexity. However, unlike inducing points, the latent variables we use do not even occupy the same space as the input data, and so offer the blessing of extra abstraction. The meta-representational aspects of the model can be related to Deep GPs, as proposed by~\citep{damianou2013deep}.

The most characteristic difference of our work to other approaches is the locality of the latent variables to units in the network, abstracting a neural network into subfunctions which are embedded in a shared structural space while still capturing dependences between weights and allowing a compact hypernetwork to not have to generate all weights of a network at once. In our case, the fact that the unit latent variables are local facilitates modeling structural dependencies in the network which leads to the ability to model weights in a more distributed fashion. This facilitates modeling adaptive subnetworks in a generative fashion as well as breaks down the immense dimensionality of weight tensors and renders the learning problem of correlated weights more tractable.

Finally, as is clearest in the permuted MNIST example, the hypernetwork can be cast as an interpreter, turning one representation of a program (the unit embeddings) into a realized method for mapping inputs to outputs (the NN). Thus, our method can be seen in terms of program induction, a field of recent interest in various fields, including concept learning~\citep{liang2010learning,perov2014learning,lake2015human,lake2018emergence}.

\section{Discussion}
\vspace{-0.1in}
We  proposed a meta-representation of neural networks. This is based on the idea of characterizing neurons in terms of pre-determined or learned latent variables, and using a shared hypernetwork to generate weights and biases from conditional distributions based on those variables.
We used this meta-representation as a  function prior, and showed its advantageous properties as a learned, adaptive, weight regularizer that can perform complexity control in function space. We also showed the complex correlation structures in the input and output weights of hidden units that arise from this meta-representation, and demonstrated how the combination of hypernetwork and network can adapt to out-of-task generalization settings and distribution shift by re-aligning the networks to the new data.
Our model handles a variety of tasks without requiring task-dependent manually-imposed structure, as it benefits from the {\it blessing of abstraction}~\citep{goodman2011learning} which arises when rich structured representations emerge from hierarchical modeling.

We believe this type of modeling jointly capturing representational uncertainty, function uncertainty and observation uncertainty can be beneficially applied to many different neural network architectures and generalized further with more interestingly structured meta-representations.

\newpage
\bibliographystyle{abbrvnat} 
\bibliography{bnn_ref}
\end{document}